\documentclass[letterpaper, 10pt, conference]{ieeeconf}

\IEEEoverridecommandlockouts
\usepackage{amsmath,amssymb,amsfonts}
\usepackage{graphicx}
\usepackage{cite}
\usepackage{booktabs}
\usepackage{url}
\usepackage{color}
\usepackage{multirow}
\usepackage{algorithm}
\usepackage{algorithmic}
\usepackage[caption=false]{subfig}
\title{\Large \bf Autonomous Droplet Navigation via Model-Based Reinforcement Learning: Zero-Shot Transfer and Emergent Dynamics
}

\author{Rajneesh Anand and Mayuresh V. Kothare
\thanks{The authors are with Department of Chemical and Biomolecular Engineering, Lehigh University, Bethlehem, PA 18015, USA {\tt raa524@lehigh.edu, mvk2@lehigh.edu}}%
}

\begin{document}

\maketitle

\begin{abstract}
Self-driving laboratories (SDLs) are transforming chemical and materials discovery through closed-loop automation, yet automated infrastructure for physical manipulation of soft, deformable matter remains beyond current robotic platforms. A critical instance is autonomous droplet transport on an open surface, where interfacial physics produce partially observable dynamics that pose significant challenges for classical model-based controllers. We introduce the first robotic platform for closed-loop autonomous liquid droplet navigation on an open, unconfined surface using model-based reinforcement learning. A two-axis tilting board coated with a thin silicone oil film drives the droplet, while an overhead camera provides real-time feedback. A learned policy was trained on just 50 to 150 physical episodes depending on geometric complexity, without simulation or analytical models. Beyond performance alone, the platform demonstrates three capabilities of interest to the SDL community: it robustly transfers zero-shot to unseen geometries; it autonomously discovers an oscillatory depinning strategy to free the droplet when it sticks; and it completes its full training pipeline in under 90 minutes. These results extend reinforcement-learning manipulation from rigid microrobots to deformable soft-matter systems for next-generation SDLs.
\end{abstract}

\section{Introduction}

Self-driving laboratories (SDLs) are transforming chemical and materials discovery by integrating robotic platforms, automated instrumentation, and machine learning into experiments~\cite{abolhasani2023rise}. 
While SDLs have automated reaction design and synthesis, automated infrastructure for physical manipulation of soft, deformable matter, where the agent has no rigid structure and the dynamics resist analytical modeling, remains beyond current robotic platforms.
A critical instance is autonomous droplet transport on an open surface. Droplet-based operations underpin lab-on-a-chip diagnostics, combinatorial chemical synthesis, and high-throughput biological assays, yet existing droplet transport platforms rely on either preprogrammed actuation sequences or manual control. The governing physics, including contact-angle hysteresis, pinning, and surface heterogeneity from lubricant depletion, introduce partially observable dynamics that pose significant challenges for classical model-based controllers.

A liquid droplet transported on an open surface is precisely such a system. Unlike rigid microrobots~\cite{medany2025modelbased}, a droplet is deformable, prone to pinning, and governed by interfacial forces~\cite{anand2024, Anand2026surfactant} that render its response to actuation nonlinear and history-dependent. Its dynamics couple contact-line motion, lubricant-film depletion, and meniscus deformation, which are computationally expensive to resolve and impractical to simulate, making  the design  of a classical model-predictive controller (MPC) challenging.

\textbf{\textit{Contributions.}} This paper demonstrates first autonomous droplet navigation on a tilt-actuated platform through open channel across five different geometries with increasing complexities. We show a zero-shot transfer learning framework in which a policy trained on the simpler (I-) shape navigates the droplet on the L-in-, Arc-in-, and a new Staircase-shape that introduces greater complexity. This study shows the first demonstration of a latent world-model agent (DreamerV3)~\cite{hafner2025mastering} trained fully offline on experimental data, without any simulator or analytical model. 

\begin{figure}[t]
\centering
\includegraphics[width=0.8\columnwidth]{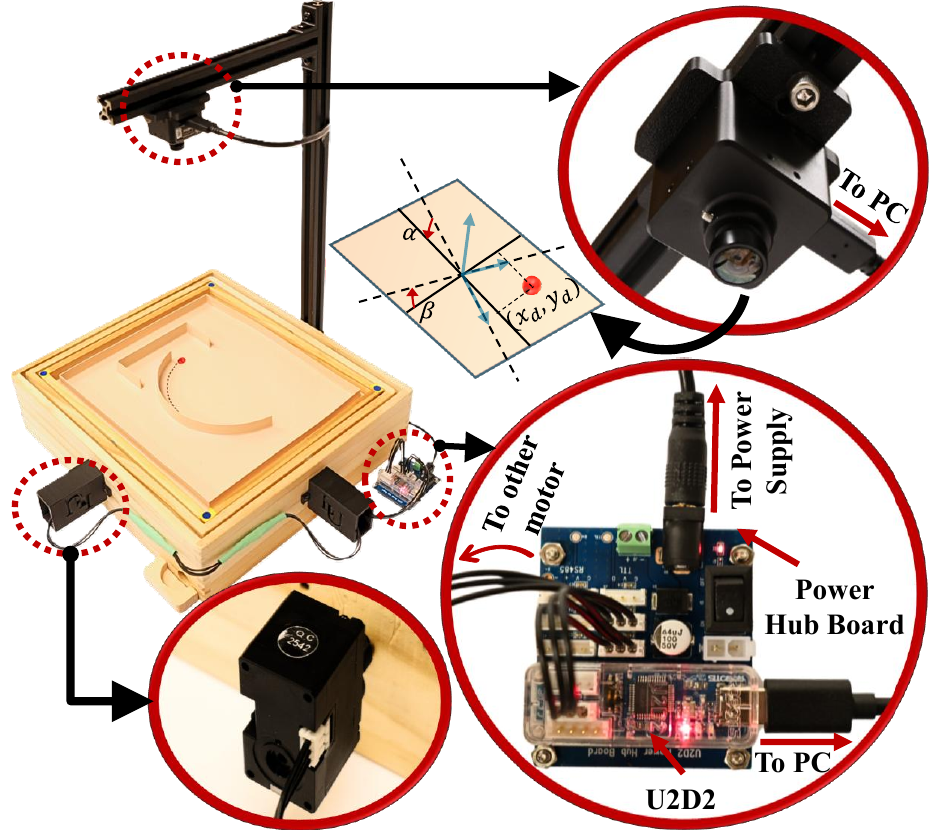}
\caption{The DropletRunner platform: 3D-printed PLA board with two-axis motor actuation, overhead camera, and U2D2 control interface.}
\label{fig:platform}
\end{figure}

\section{Methods}
The robotic (DropletRunner) platform (Fig.~\ref{fig:platform}) consists of five parts including a BRIO Labyrinth board modified with custom 3D-printed PLA channel geometries coated with a thin silicone oil film to mitigate surface problem~\cite{Anand2026surfactant}, two motors that control the inclination of the board, an overhead camera provides visual feedback, a control unit that runs all the algorithms, and finally a red-dyed water droplet. 

\subsection{Camera}

We install an overhead See3CAM$\_$24CUG camera that was mounted on T-slotted aluminum extrusion profiles with a 3D-printed mounting plate to hold the camera at a fixed height. The camera provides a 1920 $\times$ 1200 RGB image of the board and droplet at 60 fps (frames per second).

\subsection{Motors}

In order to control the inclination of the Labyrinth, both manual control knobs were removed by prying them and two Dynamixel motors were coupled to the exposed tilt shafts. Pilot holes were drilled into the Labyrinth frame, and 3D-printed motor housings were slid over the motors and bolted to the frame to fix the motors in place. Both motors were daisy-chained using a ROBOT X3P cable extended by splicing additional wire between the corresponding pins. The motor chain was connected to a U2D2 USB-to-Dynamixel adapter mounted on a U2D2 Power Hub Board, which was powered by a 5V3A USB supply. The U2D2 and camera were connected to the host laptop via USB. A complete closed-loop hardware chain is shown in Fig.~\ref{fig:platform}.

\subsection{Control Unit}
The control unit uses two computers: a CPU-only laptop that runs the ROS2 control loop and motor communication via the U2D2 adapter, and a separate GPU platform used for offline world-model training. Collected episodes are transferred to the GPU for training, and the resulting policy is sent back for evaluation. This decoupled setup prevents training latency from interfering with real-time control.

\section{Results}
\begin{figure}[t]
\centering
\includegraphics[width=0.8\columnwidth]{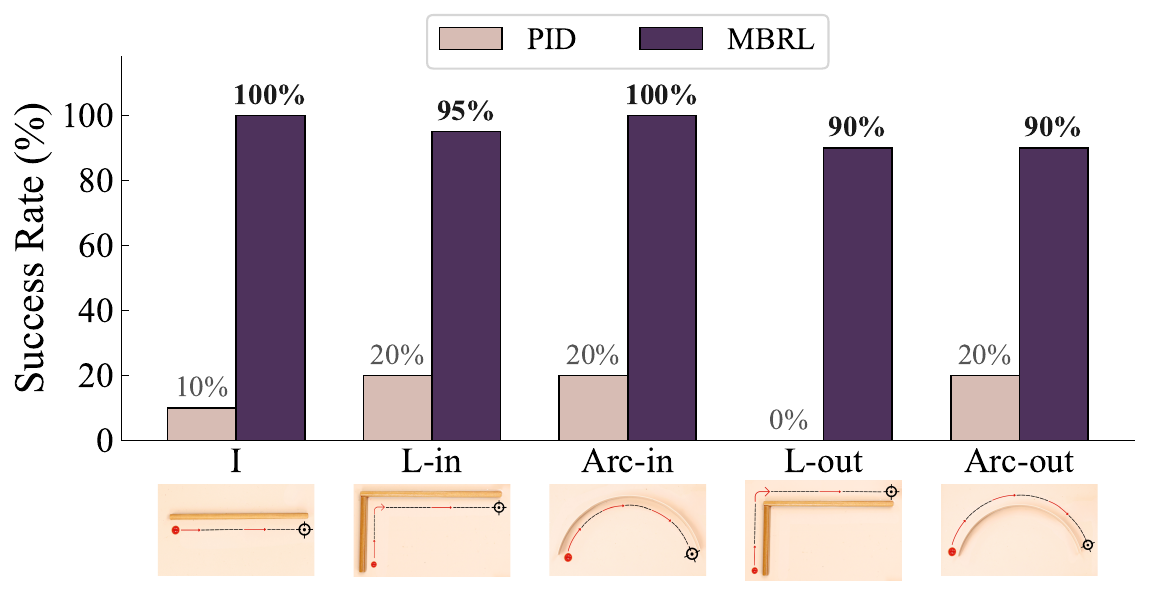}
\caption{Success rate to achieve the desired target for PID and MBRL across five geometries shown below.}
\label{fig:successrate}
\end{figure}

We present DropletRunner\footnote{The source code for autonomous droplet naviagtion is available at \url{https://github.com/rajneeshanand/DropletRunner.}}~\cite{anand2026dropletrunner}, to our knowledge the first robotic platform for closed-loop autonomous liquid droplet navigation on an open, unconfined surface, enabled by model-based reinforcement learning (MBRL). The platform (Fig.~\ref{fig:platform}) consists of a 3D-printed PLA board coated with a thin silicone oil film of unobservable thickness, actuated by two motors that provide two-axis gravitational tilting, analogous to the classic Labyrinth platform. An overhead camera provides real-time visual feedback, and the complete perception-action loop runs at 20\,Hz on a standard laptop. To date, this is the first demonstration of a latent world-model agent (DreamerV3)~\cite{hafner2025mastering} trained fully offline on experimental data, without any simulator or analytical model. We demonstrate autonomous navigation across five channel geometries of increasing complexity: straight (I-shape), right-angle L-shapes (L-inside, L-outside), and continuous-curvature arcs (Arc-inside, Arc-outside). The MBRL policy achieves 100\% success on straight paths, 95\% on L-inside, 100\% on Arc-inside, and 90\% on unconfined exterior-wall geometries (L-outside, Arc-outside), shown in Fig.~\ref{fig:successrate}. An optimized PID baseline on the same hardware achieves approximately 20\%. An open-loop ablation confirms the policy actively corrects in real time rather than replaying a memorized sequence.

We highlight three findings relevant to the broader SDL community. First, the agent autonomously discovers an oscillatory depinning strategy at geometric constrictions, alternating-polarity tilt commands that escalate when the droplet sticks. This emergent behavior, never reward-shaped or programmed, is consistent with vibration-induced depinning, demonstrating that MBRL can rediscover physical transport mechanisms. Second, an I-shape policy achieves 95\% zero-shot success on the L-shape and Stair-shape. This policy also achieves 100\% on the Arc after only 20 fine-tuning episodes, a five-fold reduction over training from scratch. Third, the entire training pipeline, from first random episode to deployed policy, requires under 90 minutes (and further reduced with transfer learning) of physical experimentation, establishing a practical template for data-scarce or simulation-free SDL subsystems.

Our system bridges the gap between AI-driven experimental design and the physical sample transport that SDLs require. To our knowledge, this is the first autonomous robotic control of a deformable liquid agent, broadening the scope of RL-based manipulation from rigid-body microrobots to soft-matter systems. We will present experimental videos and discuss how learned policies can complement or replace hand-crafted controllers for autonomous manipulation of partially observable systems in next-generation SDL workflows.

\section{Conclusion}

In this work, we present an efficient approach to enable autonomy in droplet navigation through complex geometries using deep RL. Readers can read the full paper~\cite{anand2026dropletrunner} for the detailed methods, results, and analysis. 
Looking further ahead, we aim to open-source the hardware and software of the robotic system, giving other researchers a testbed that uses gravity-driven oil drag for droplet navigation. We believe it offers a strong real-world benchmark with low space requirements, low hardware complexity, automated lab equipment and modest cost.

\bibliographystyle{IEEEtran}
\bibliography{citation}

@article{abolhasani2023rise,
  author  = {Abolhasani, Milad and Kumacheva, Eugenia},
  title   = {The rise of self-driving labs in chemical and materials sciences},
  journal = {Nature Synthesis},
  volume  = {2},
  pages   = {483--492},
  year    = {2023},
  doi     = {10.1038/s44160-022-00231-0}
}

@article{hafner2025mastering,
  author  = {Hafner, Danijar and Pasukonis, Jurgis and Ba, Jimmy and Lillicrap, Timothy},
  title   = {Mastering diverse domains through world models},
  journal = {Nature},
  volume  = {640},
  pages   = {647--653},
  year    = {2025},
  doi     = {10.1038/s41586-025-08744-2}
}

@article{anand2026dropletrunner,
  author  = {Anand, Rajneesh and Kothare, Mayuresh V.},
  title        = {Autonomous Droplet Navigation via Model-Based Reinforcement Learning},
  journal      = {arXiv preprint arXiv:2609.16369},
  year         = {2026},
  archivePrefix= {arXiv},
  primaryClass = {cs.LG},
}

@article{anand2024,
 author={Anand, Rajneesh and Mandal, Mrinal K. and Chaudhuri, Rajib G. and Hens, Abhiram},
 title = {Effect of patterned condensation surface on the performance of a newly designed basin-type solar desalination unit},
 journal = {Applied Thermal Engineering},
 volume = {249},
 pages = {123399},
 year = {2024},
}

@article{Anand2026surfactant,
  author  = {Anand, Rajneesh and Kothare, Mayuresh V.},
  title   = {Surfactant-free splitting of water droplets in a microchannel},
  journal = {Analyst},
  volume={151},
  pages={5095--5106},
  year    = {2026},
  doi     = {10.1039/d6an00516k},
}

@article{medany2025modelbased,
  author  = {Medany, M. and Piglia, L. and Achenbach, L. and Mukkavilli, S. K. and Ahmed, D.},
  title   = {Model-based reinforcement learning for ultrasound-driven autonomous microrobots},
  journal = {Nat. Mach. Intell.},
  volume  = {7},
  pages   = {1076--1090},
  year    = {2025}
}

\end{document}